\documentclass[runningheads]{llncs}
\usepackage[T1]{fontenc}

\usepackage{microtype}
\usepackage{graphicx}
\usepackage{booktabs} 
\usepackage{subcaption}

\usepackage{hyperref}
\usepackage{wrapfig, lipsum}
\usepackage[utf8]{inputenc}
\usepackage{microtype}
\usepackage{graphicx}
\usepackage{booktabs} 
\usepackage{url}            
\usepackage{nicefrac}       
\usepackage{xspace}
\usepackage{enumerate}
\usepackage{amsmath, mathtools, amsfonts, amssymb, dsfont, enumitem, bm, bbm, mathabx}
\usepackage{graphicx}
\usepackage[mathscr]{euscript}
\usepackage[capitalize,noabbrev]{cleveref}
\usepackage{adjustbox}
\usepackage{footmisc}
\usepackage[skip=5pt]{caption}
\usepackage[skip=3pt]{subcaption}

\usepackage{soul}
\usepackage[capitalize,noabbrev]{cleveref}
\usepackage{floatrow}

\usepackage{booktabs}
\usepackage{multirow}

\usepackage{amsmath}
\usepackage{amssymb}
\usepackage{mathtools}

\usepackage[capitalize,noabbrev]{cleveref}

\usepackage[textsize=tiny]{todonotes}

\usepackage{placeins}
\usepackage[utf8]{inputenc}
\usepackage{colortbl}
\usepackage{url}
\usepackage{amsmath} 
\usepackage{amssymb}  
\usepackage{graphicx}
\usepackage{hyperref}
\usepackage{mathtools}
\usepackage{xcolor}
\usepackage{bm}
\usepackage{graphicx}
\usepackage{siunitx}
\usepackage{tabularx}

\newcommand{\eps}{\varepsilon}

\renewcommand{\bar}{\widebar}

\newcommand{\R}{\mathbb{R}}

\newcommand{\N}{\mathcal{N}}

\usepackage{graphicx}
\begin{document}
\title{Aligning Diffusion Model with Problem Constraints for Trajectory Optimization}

%
%
\author{Anjian Li\inst{1}
\and Ryne Beeson\inst{2}}
\authorrunning{A. Li et al.}
%
\hypersetup{
    colorlinks=true,
    linkcolor=blue,
    urlcolor=blue,
    citecolor=blue,
    pdfborder={0 0 0}
}
\institute{Department of Electrical and Computer Engineering, Princeton University 
\and Department of Mechanical and Aerospace Engineering, Princeton University 
\\
\email{\{anjianl, ryne\}@princeton.edu}}
\maketitle              
\begin{abstract}

Diffusion models have recently emerged as effective generative frameworks for trajectory optimization, capable of producing high-quality and diverse solutions. 
However, training these models in a purely data-driven manner without explicit incorporation of constraint information often leads to violations of critical constraints, such as goal-reaching, collision avoidance, and adherence to system dynamics. 
To address this limitation, we propose a novel approach that aligns diffusion models explicitly with problem-specific constraints, drawing insights from the Dynamic Data-driven Application Systems (DDDAS) framework.
Our approach introduces a hybrid loss function that explicitly measures and penalizes constraint violations during training. 
Furthermore, by statistically analyzing how constraint violations evolve throughout the diffusion steps, we develop a re-weighting strategy that aligns predicted violations to ground truth statistics at each diffusion step.
Evaluated on a tabletop manipulation and a two-car reach-avoid problem, our constraint-aligned diffusion model significantly reduces constraint violations compared to traditional diffusion models, while maintaining the quality of trajectory solutions.
This approach is well-suited for integration into the DDDAS framework for efficient online trajectory adaptation as new environmental data becomes available.

\end{abstract}

\section{Introduction}

\begin{figure}
    \centering
    \includegraphics[width=0.8\linewidth]{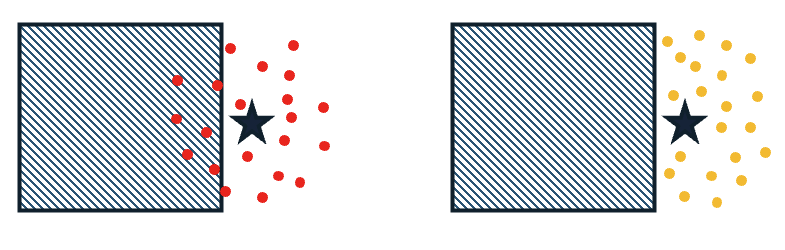}
    \caption{Left: Unconstrained diffusion samples (red) around a local optima (star) with an infeasible region (grey) presented. Right: Constraint-aligned diffusion samples (yellow) around the same local optima.}
    \label{fig:add constraints}
\end{figure}

Trajectory optimization involves finding optimal trajectories to reach specific goals while minimizing or maximizing certain performance criteria, such as time, energy consumption, or fuel usage, subject to various system constraints.
Such constrained optimization problems typically demand trajectories that closely follow system dynamics, satisfy boundary conditions (start and goal configurations), and respect safety and operational constraints.
Importantly, in many constrained optimization scenarios, the optimal solutions often lie on the boundaries of constraints, or even at intersections of multiple constraint boundaries. 
Thus, methods capable of efficiently generating high-quality feasible solutions positioned close to constraint boundaries---without violating them---are crucial, shown in Fig. \ref{fig:add constraints}, as these solutions can either be used directly online or leveraged as initial guesses to accelerate subsequent optimization processes.

An example of constrained trajectory optimization arises in autonomous robot navigation, where a robot must navigate from a predefined start position to a designated goal location. 
Here, planned trajectories must closely follow the robot's nonlinear system dynamics and find collision-free paths in complex, obstacle-rich environments.
Constraint satisfaction, especially regarding obstacle avoidance, is important in these safety-critical systems, where constraint violations could lead to severe damage for both the robot and its surroundings.

Another relevant application is spaceflight trajectory optimization, where trajectories are highly nonlinear and governed by complex gravitational dynamics. Such problems often lead to high-dimensional, non-convex optimization landscapes, exhibiting multiple local optima. 
In this case, identifying a collection of diverse, locally optimal solutions is often valuable, as tradeoffs can be made across various objectives.
This requires a global search algorithm.
Traditionally, evolutionary algorithms \cite{mitchell1998introduction,kennedy1995particle} and Monotonic Basin Hopping (MBH) \cite{wales1997global,leary2000global} methods have been employed to explore these global solution spaces \cite{izzo2007search,vinko2007benchmarking,yam2011low}. However, these approaches are computationally intensive and rarely leverage solution structures from similar problems in the search process.

\begin{figure}
    \centering
    \includegraphics[width=1.0\linewidth]{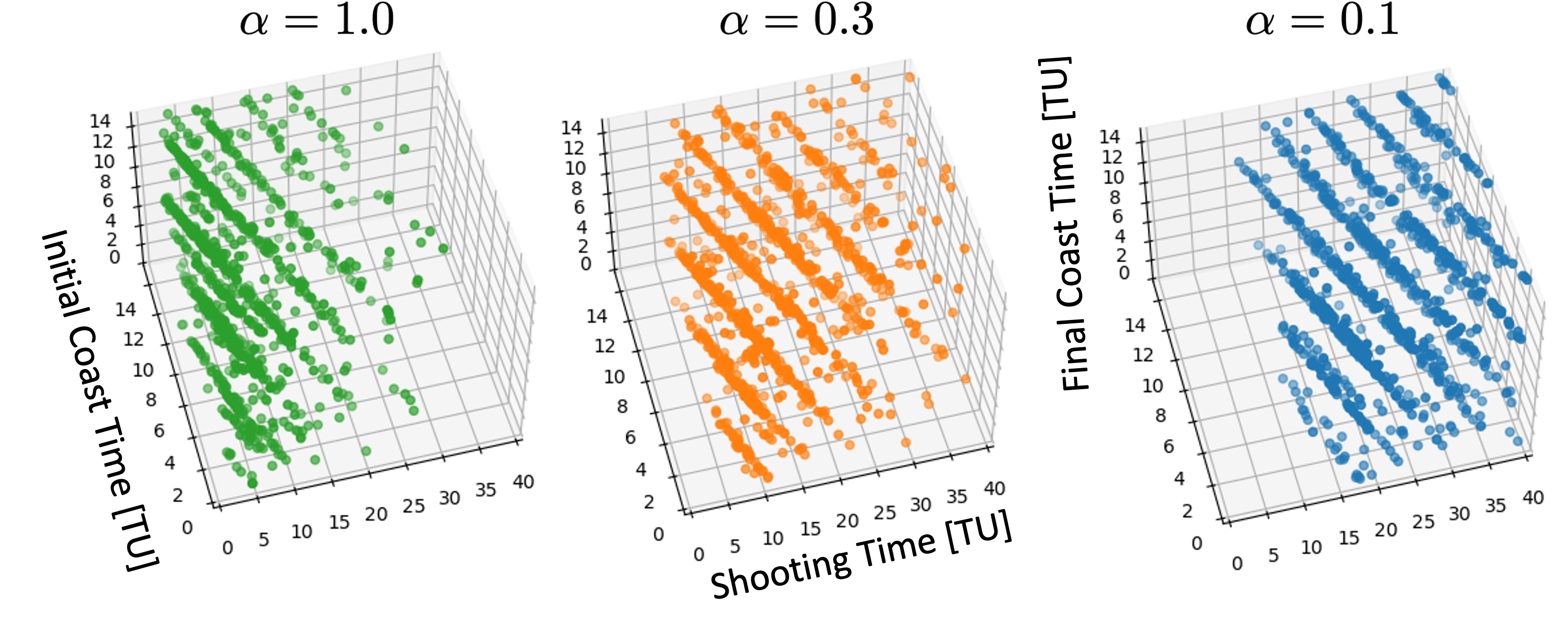}
    \caption{Local optimal solution structure varies with optimization problem parameter $\alpha$. Figure is adopted from \cite{beeson2024global}.}
    \label{fig: solution structure}
\end{figure}

It has been widely recognized that similar optimization problems often exhibit consistent solution patterns or structures \cite{amos2023tutorial}. 
Recent studies in trajectory optimization explicitly identify such structured characteristics \cite{beeson2024global,graebner2024learning}. 
For instance, Fig. \ref{fig: solution structure} from \cite{beeson2024global} illustrates structured solution patterns for a low-thrust cislunar transfer problem. 
The problem parameter $\alpha$ represents the maximum allowable thrust of the spacecraft, and each scenario (with $\alpha=0.1, 0.3, 1.0$) is highly non-convex and features multiple local minima.
The Fig. \ref{fig: solution structure} highlights structured solutions for time variables---initial and final coast times and shooting times---grouped into parallel hyperplanes. 
These structured patterns shift systematically as the parameter $\alpha$ changes. 
Learning these underlying structural variations can significantly accelerate the generation of solutions for new, previously unseen parameter values.

Motivated by these observations, the Amortized Global Search (AmorGS) method \cite{li2023amortized,beeson2024global} was recently introduced to exploit machine learning techniques for accelerating global searches in trajectory optimization problems. 
The key insight is to amortize the computation burden offline: by solving many similar problem instances in advance, we leverage the knowledge of these problems to accelerate the global search for new problem instances online.
A crucial aspect of AmorGS is the careful selection of generative machine learning architectures that effectively capture the problem-specific solutions structure with good generalizations \cite{li2023amortized,beeson2024global}.

Diffusion models \cite{sohl2015deep,ho2020denoising,song2020score} have recently demonstrated promising performance in generating high-quality samples for complex data, including images \cite{rombach2022high}, videos \cite{ho2022video}, and robot trajectories \cite{janner2022planning,ajay2022conditional,chi2023diffusion,zhang2024predicting}, based on pre-collected datasets.
Within the AmorGS framework, diffusion models have proven particularly successful, efficiently sampling near-optimal and diverse trajectory candidates and significantly outperforming other generative architectures like variational auto-encoders (VAEs) \cite{li2024diffusolve}. 
Additionally, diffusion models exhibit strong generalization capabilities, effectively adapting to problem environments and parameters not encountered during training.

However, despite learning to sample from the distribution of locally optimal trajectories, diffusion models may still violate critical constraints (e.g., goal-reaching, collision avoidance, and system dynamics), as they learn only from optimal data without explicitly considering the problem's constraint information, as illustrated in Fig. \ref{fig:add constraints}. 
Satisfying these constraints is crucial for effective and safe robot navigation.
Fortunately, unlike black-box optimization problems commonly encountered in biology, chemistry, or material science \cite{trabucco2022design,krishnamoorthy2023diffusion}, trajectory optimization constraints often have clearly defined analytical forms. 
For example, constraints can be explicitly represented, such as requiring the minimum distance between the robot trajectory and obstacles to exceed a safety threshold. 
Nevertheless, encoding such explicit constraint information into diffusion models remains challenging because diffusion models aim to learn the error term between two diffusion steps during the denoising process.
As a result, directly evaluating constraint violations from diffusion model outputs is difficult. 
Furthermore, when using standard diffusion methods with Gaussian noise, any bounded constraint will have a nonzero probability of being violated due to perturbations in the noisy samples. 
Consequently, even minor perturbations from the tails of Gaussian distributions during the diffusion process can lead to constraint violations.

To address these challenges, we propose a novel approach that explicitly aligns diffusion models with constraint information, enabling the generation of high-quality solutions with improved constraint satisfaction compared to traditional diffusion models.
During training, we utilize the error term predicted by the diffusion model to perform one step of reverse sampling, enabling direct evaluation of constraint violation loss from the current diffusion model output. 
Additionally, we analyze how constraint violations statistically increase as we corrupt ground truth training data, enabling us to develop an appropriate re-weighting scheme aligned with the statistical understanding of ground truth constraint violations at each diffusion step.
With these improvements, we train our constraint-aligned diffusion model using a novel hybrid loss function that combines the original diffusion model loss with a re-weighted constraint violation loss.

\begin{figure}[tp]
    \centering
    \includegraphics[width=0.7\textwidth]{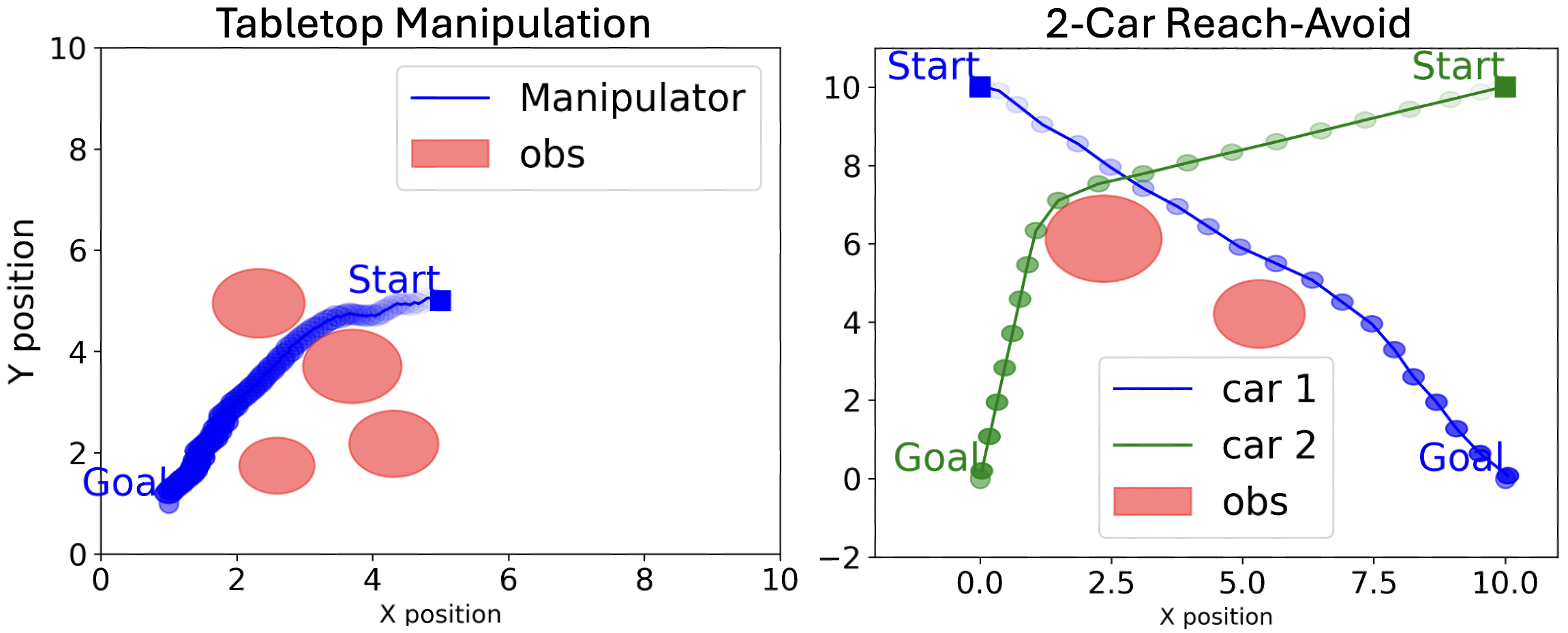}
    \caption{Example trajectory for the tabletop manipulation and 2-car reach-avoid. ``Obs" are short for obstacles.
    }
    \label{fig: example trajectory}
\end{figure}

To demonstrate the effectiveness of our approach, we apply our constraint-aligned diffusion model to two non-convex trajectory optimization problems: a tabletop manipulation problem and a two-car reach-avoid problem, as shown in Fig. \ref{fig: example trajectory}.
First, we examine how constraint violation statistics evolve as we corrupt the locally optimal trajectory in the training data for these two problems.
Then, we demonstrate that by incorporating the re-weighted constraint violation loss, our model generates significantly more feasible trajectories compared to traditional unconstrained diffusion models.

Our proposed approach utilizes core principles of the Dynamic Data Driven Applications Systems (DDDAS) framework by leveraging pre-collected data to train conditional diffusion models that can be applied to new problem instances on the fly. 
This research establishes a foundation for the integration of efficient generative models within DDDAS applications that can efficiently generate feasible robot trajectories in real time based on streaming sensor data. 
In the future, data collected during online deployment could be used to continuously refine the diffusion model, creating a feedback loop that enhances constraint satisfaction over time and enabling robots to adapt to changing environments with this dynamic data-driven approach.

\section{Related Work}

\paragraph{Amortized Optimization}
Machine learning has been widely adopted to accelerate solving optimization problems through amortization \cite{amos2023tutorial}.
For example, multilayer perceptions (MLPs) have been adopted to predict solutions to Quadratic Programming (QP) problems \cite{zhang2019safe,sambharya2024learning}.
Additionally, machine learning models have been utilized to learn strategies that speed up finding solutions for mixed-integer optimization \cite{bertsimas2021voice} and combinatorial optimization \cite{sun2023difusco}.
In particular, Amortized Global Search (AmorGS) \cite{li2023amortized,beeson2024global} has proposed a generative model approach for solving non-convex trajectory optimization problems by leveraging solution structures learned from similar problems. 
This framework has been further enhanced through the incorporation of diffusion models \cite{li2024diffusolve} with performance improvement.

\paragraph{Diffusion Models.}
Diffusion models \cite{sohl2015deep,ho2020denoising,song2020score} are powerful generative models that sample from a complex data distribution through a forward diffusion process and a learnable reverse denoising process.
These models have demonstrated remarkable success in generating high-quality images and videos \cite{rombach2022high,ho2022video}, biological structures such as proteins and molecules \cite{hoogeboom2022equivariant,ketata2023diffdock}, and robot trajectories and control sequences  \cite{janner2022planning,ajay2022conditional,chi2023diffusion,li2024diffusolve,li2025end}.
The diffusion model also allows the incorporation of contextual information for conditional generation through various approaches, including classifier guidance \cite{dhariwal2021diffusion} or classifier-free guidance \cite{ho2022classifier}.
These capabilities make diffusion models particularly well-suited for generating high-quality trajectory samples from distributions conditioned on specific parameters of trajectory optimization problems.

\paragraph{Diffusion Model with Constraints.}
Across different domains, researchers have developed various methods to incorporate constraints into diffusion models, primarily focusing on the sampling process.
In topology optimization, diffusion sampling is guided by structural preferences through classifier guidance techniques \cite{maze2023diffusion}.
Similarly, for robot trajectory generation, control barrier functions are employed during sampling to enhance safety constraint satisfaction \cite{botteghi2023trajectory}. 
Projected diffusion models take a different approach to keep samples within a feasible region through alternating projection and denoising steps \cite{christopher2024constrained}. 
For inverse problems like image inpainting, researchers have developed specific correction terms to constrain the diffusion process within certain data manifolds \cite{chung2022improving}. 
When handling multiple constraints simultaneously, compositional techniques have proven effective during the sampling phase \cite{yang2023compositional,power2023sampling}.

Our approach differs from existing work in two fundamental ways.
First, we integrate constraints directly into the diffusion model training process, which allows our model to combine with existing methods during sampling to further improve the performance.
Second, we introduce a novel re-weighting scheme based on a statistical understanding of how constraint violations naturally evolve throughout the diffusion process. 
This principled approach enables more effective constraint satisfaction across all diffusion steps.
\section{Preliminaries}

In this section, we introduce the trajectory optimization problem and the original unconstrained diffusion model with classifier-free guidance for conditional generation \cite{ho2020denoising,ho2022classifier}.
We then briefly discuss how diffusion models can be integrated into the Amortized Global Search (AmorGS) framework to accelerate the problem-solving process.

\subsection{Trajectory Optimization Problem}
In this paper, we aim to solve a trajectory optimization problem formulated as a parameterized Nonlinear Program (NLP):
\begin{align} \label{eq: trajectory optimization prob}
    \mathcal{P}_{y} \coloneqq 
    \begin{cases}
    \underset{x}{\min} \quad &J(x;y)  \\
    s.t., \quad &g_{i}(x;y) \leq 0, \ i = 1, 2, ..., l \\
    \quad &h_{j}(x;y) = 0, \ j = 1, 2, ..., m  \\
    \end{cases}  
\end{align}
where $J \in \mathcal{C}^1(\R^n, \R^k; \R)$ represents the objective function to minimize, such as the time to reach the goal or fuel expenditure.
$g_i \in \mathcal{C}^1(\R^n, \R^k; \R)$ and $h_j \in \mathcal{C}^1(\R^n, \R^k; \R)$ are both constraint functions that include dynamic constraints, collision avoidance, goal-reaching, etc.
The optimization variable $x \in \R^n$ can represent robot controls.
$y \in \R^k$ denotes problem parameters such as goal positions, obstacle locations, control limits, etc., that vary with changing environments or tasks.

\subsection{Unconstrained Diffusion Model}\label{sec: original diffusion model}

We adopt the Denoising Diffusion Probabilistic Model (DDPM) \cite{ho2020denoising} as our unconstrained baseline for trajectory generation. 
The diffusion model consists of a forward (noising) process and a reverse (denoising) process.

In the forward process, Gaussian noise is gradually added to the clean data sample \(x_0\) over \(K\) steps. At each step \(k\), the noisy sample \(x_k\) is obtained by:
\begin{align} \label{eq: forward sample}
    x_k = \sqrt{\bar{\alpha}_k} \, x_0 + \sqrt{1 - \bar{\alpha}_k} \, \eps, \quad \eps \sim \mathcal{N}(0, I), \quad 0 \leq k \leq K,
\end{align}
where \(\bar{\alpha}_k\) is a predefined noise schedule that determines the level of noise added at step \(k\).

The reverse process aims to recover the original data by predicting the added noise. Given a condition variable \(y\), the baseline model follows the classifier-free guidance approach \cite{ho2022classifier}, training the model to predict both conditional noise $\eps_\theta(x_k, k, y)$ and unconditional noise $\eps_\theta(x_k, k, y = \varnothing)$. 
With probability \(p_{\text{uncond}}\), the model is trained without conditioning. Letting \(b \sim \text{Bernoulli}(p_{\text{uncond}})\), the training objective becomes:
\begin{align} \label{eq: diffusion model loss}
    \mathcal{L}_\text{diff} =
    \mathbb{E}_{(x_0, y), k, \eps, b} \left\| \eps_\theta\Big(x_k(x_0, \eps), k, (1 - b) \cdot y + b \cdot \varnothing\Big) - \eps \right\|_2^2.
\end{align}

During sampling, the model generates data by iterative denoising from \(x_K\) to \(x_0\). The reverse step from \(x_k\) to \(x_{k-1}\) is given by:
\begin{align} \label{eq: reverse sample}
    x_{k-1} = \frac{1}{\sqrt{\alpha_k}} \left( x_k - \frac{\beta_k}{\sqrt{1 - \bar{\alpha}_k}} \, \bar{\eps}_\theta(x_k, k, y) \right) + \sqrt{\beta_k} z,
\end{align}
where \(z \sim \mathcal{N}(0, I)\), \(\alpha_k = 1 - \beta_k\), and \(\beta_k\) is the noise variance schedule.
The guided noise predictor \(\bar{\eps}_\theta\) is defined as:
\begin{align}
    \bar{\eps}_\theta(x_k, k, y) = (\omega + 1) \eps_\theta(x_k, k, y) - \omega \eps_\theta(x_k, k, y = \varnothing),
\end{align}
where \(\omega\) controls the weight of the conditional generation.

\subsection{AmorGS with Diffusion Models} \label{sec: amorgs with diffusion}

In the AmorGS framework \cite{li2023amortized,beeson2024global}, we first create diverse problem instances $\mathcal{P}_{y}$ through uniform sampling of parameter $y$ within a reasonable range.
Next, we use Sparse Nonlinear OPTimizer (SNOPT) \cite{gill2005snopt}, a sequential quadratic programming (SQP) based solver \cite{boggs1995sequential}, to obtain a collection of diverse and locally optimal solutions to each problem $\mathcal{P}_{y}$ in Eq. \eqref{eq: trajectory optimization prob}.
These solutions and their corresponding problem parameters are paired to form our training dataset $(x^*, y)$.
To ensure data quality, we set some objective thresholds that filter out suboptimal solutions. 
Finally, we train a diffusion model to sample high-quality trajectory solutions $x^*$ conditioned on problem parameter $y$, hopefully generalizing to new problem instances with previously unseen parameter values.
\section{Methodology}

\subsection{Importance of Constraints Information}

As discussed in Sec. \ref{sec: original diffusion model} and Sec. \ref{sec: amorgs with diffusion}, within the AmorGS framework, the original unconstrained diffusion model learns from the training data and sample trajectory solutions conditioned on problem parameters. 
Since the training data consists of locally optimal (and thus feasible) solutions collected by the SNOPT solver, one might expect diffusion samples to also be nearly optimal and feasible. 
However, this purely data-driven approach has fundamental limitations.

Consider a common scenario in trajectory optimization problems illustrated in Fig. \ref{fig:add constraints}, where an infeasible region exists near a locally optimal solution—for example, a time-optimal path might pass very close to an obstacle to minimize travel time. 
When the diffusion model is trained solely on locally optimal solutions using the loss function in Eq. \eqref{eq: diffusion model loss}, it remains unaware of constraint information encoded in functions $g$ and $h$ from Eq. \eqref{eq: trajectory optimization prob}. 
Due to the expressiveness of the diffusion model and inevitable prediction errors during generalization, samples typically distribute around the true local optimum, resulting in some overlap with the infeasible region, as shown on the left of Fig. \ref{fig:add constraints}.

This observation motivates our design of a constraint-aligned diffusion model that leverages problem structure, particularly constraint functions, during the training process. 
Our goal is to shift the sampling distribution toward the feasible region (as depicted on the right of Fig. \ref{fig:add constraints}), thereby reducing constraint violations in the generated samples.

\begin{figure}
    \centering
    \includegraphics[width=0.8\linewidth]{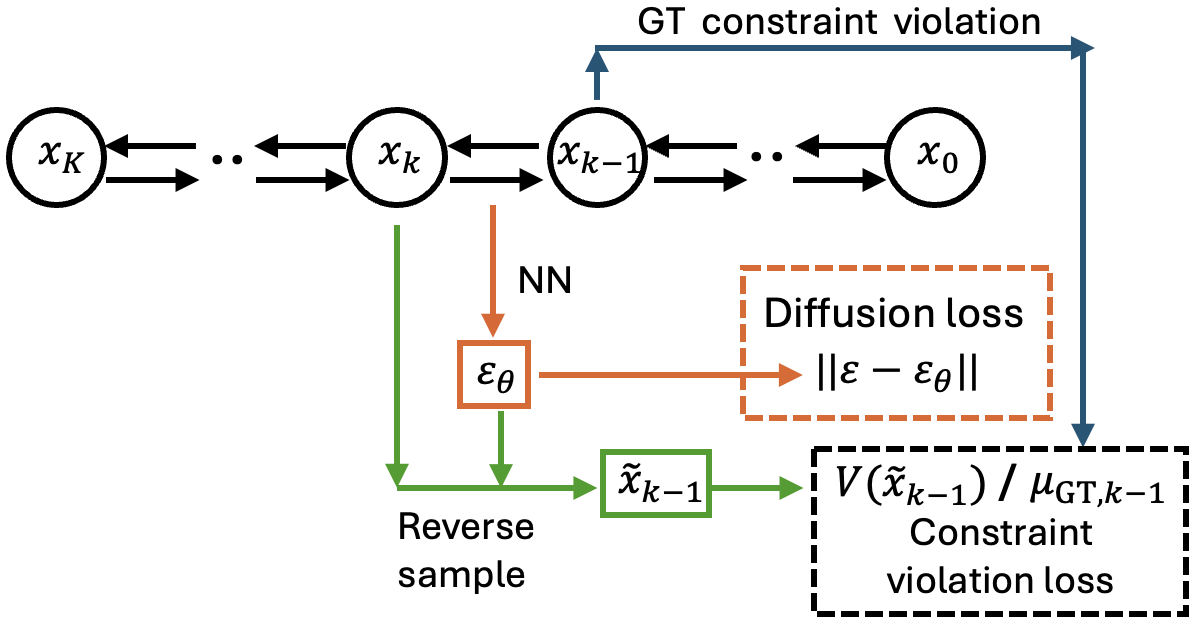}
    \caption{Hybrid training loss computation in constraint-aligned diffusion models. ``GT'' denotes ground truth. ``NN" denotes neural networks.}
    \label{fig: constraint violation loss computation}
\end{figure}

\subsection{Constraint-Aligned Diffusion Model}

In this section, we present a novel constraint-aligned diffusion model that samples high-quality solutions while aligning with problem constraint information with fewer constraint violations.
It is achieved with a novel hybrid loss function during training.
The computation workflow of this loss function, including the diffusion loss and constraint violation loss, is summarized in Fig. \ref{fig: constraint violation loss computation}.

We first define $V(x;y) \in \mathcal{C}^1(\R^n, \R^k;\R)$ as the sum of all constraint violation values from Eq. \eqref{eq: trajectory optimization prob}:
\begin{align}
    V(x, y) &= \sum_{i=1}^l \max(g_i(x;y), 0) + \sum_{j=1}^m  |h_j(x;y)| \nonumber
\end{align}

Since $V$ evaluates on data $x_k$ and $y$ instead of $\eps_\theta(x_k, k, y)$ from the neural network, we perform a one-step reverse sampling to predict $\tilde x_{k-1}$ from $x_k$ similar to \eqref{eq: reverse sample} but with only conditional noise $\eps_\theta(x_k, k, y)$, shown in Fig. \ref{fig: constraint violation loss computation}.
Since the data $\tilde x_k$ is noisy,  we clip $\tilde x_k$ to have the same range as $x_0$.
Now we are able to evaluate the violation $V(\tilde x_{k-1}, y)$, given $x_0, y, k$ and $\eps_\theta$, where the gradient information will be back-propagated through $\eps_\theta$.
We then introduce the violation loss $\mathcal{L}_\text{vio}(x_0, y, k)$:
\begin{align} \label{eq: predicted violation loss}
     \mathcal{L}_\text{vio}(x_0, y, k) &= \mathbb{E}_{\eps, z}\bigg[ V\Big(\tilde x_{k-1}(x_k(x_0, \eps), \eps_\theta, z), y\Big)\bigg]
\end{align}

However, $V(\cdot)$ is not expected to be zero on the approximated noisy data $\tilde x_{k-1}$.
Because even for $x_{k-1}$ corrupted from ground truth data, it has certain constraint violations depending on the noise level. 
To understand the ground truth constraint violation in the intermediate diffusion steps, we sample $x_{k-1}$ from $\N( \sqrt{\bar{\alpha}_{k-1}} x_0, \sqrt{1 - \bar{\alpha}_{k-1}}I)$ offline for $N$ times,
and compute the average violation value $\mathcal{\mu}_{\text{vio\_GT}}$ as the approximation of the ground truth violation:
\begin{align} \label{eq: groundtruth violation loss mean}
    &\mathcal{\mu}_{\text{vio\_GT}} (x_0, y, k) = \mathbb{E}_{\eps} \bigg[ V\Big( x_{k-1}(x_0, \eps), y\Big)\bigg] 
\end{align}

We then use this ground truth violation $\mathcal{\mu}_{\text{vio\_GT}}$ in Eq. \eqref{eq: groundtruth violation loss mean} to re-weight the constraint violation loss $\mathcal{L}_\text{vio}$ in Eq. \eqref{eq: predicted violation loss} at each diffusion step. 
Finally, we introduce a hybrid loss function from Eq. \eqref{eq: diffusion model loss}, \eqref{eq: predicted violation loss}, \eqref{eq: groundtruth violation loss mean}:
\begin{align} \label{eq: hybrid loss}
    \mathcal{L_{\text{constrained\_diff}}} &= \mathcal{L}_\text{diff} + \lambda \cdot \frac{\mathcal{L}_\text{vio}}{\mathcal{\mu}_{\text{vio\_GT}}}
\end{align}
The key insight is that the constraint violation loss $\mathcal{L}_\text{vio}$ receives less penalty when the ground truth violation $\mathcal{\mu}_{\text{vio\_GT}}$ is large.
This indicates that the constraint violation carries more weight in the overall loss function when the diffusion step $k$ is small and the data contains less noise.
\section{Experiment}

\begin{figure}[tp]
    \centering
    \includegraphics[width=0.9\textwidth]{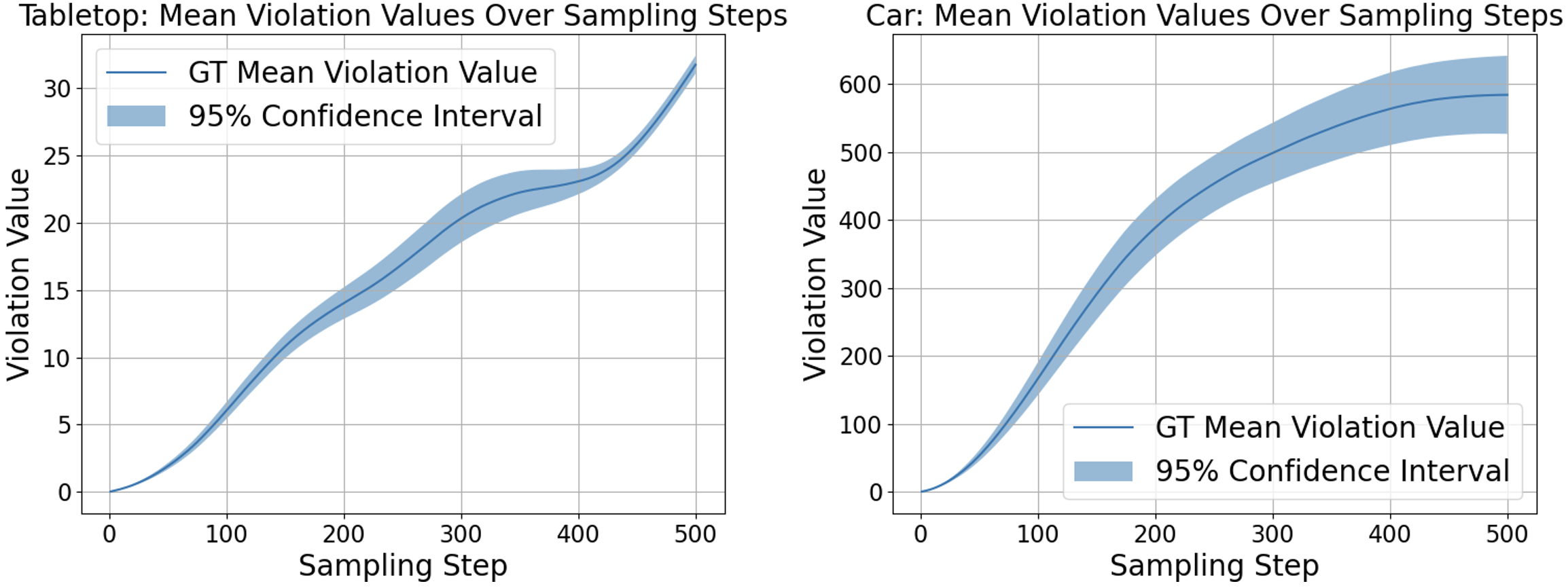}
    \caption{Ground truth constraint violation. Left: Tabletop Manipulation. Right: Two-car Reach-Avoid.}
    \label{fig: tabletop car gt violation}
\end{figure}

We validate our proposed approach using two trajectory optimization problems: a tabletop manipulation problem and a two-car reach-avoid problem, as depicted in Fig. \ref{fig: example trajectory}.
We first introduce the problem and training setup.
Then, we study how ground truth constraint violation statistics evolve as we corrupt the training data.
Finally, we compare the trajectory samples of our constraint-aligned diffusion model with the unconstrained diffusion model, showing the superiority of reducing constraint violation of our methods while preserving the sample quality.

\subsection{Problem Setup}

\subsubsection{Tabletop Manipulation}

In this problem, the objective is to minimize the total time $t$ required for the end effector of a manipulator to reach a table corner while avoiding collisions with obstacles, as illustrated on the left of Fig.~\ref{fig: example trajectory}. The manipulator always starts from the center of the table.

The problem parameters are denoted by $ y = (p_{\text{goal}}, p_{\text{obs}}, r_{\text{obs}}) $, where $ p_{\text{goal}} $ represents the goal location, sampled randomly from one of the four table corners. 
Four obstacles are placed randomly between the start and goal positions, with at least one obstacle placed between the start and goal.
The obstacle radii $r_{\text{obs}}$ are sampled from $[0.3, 1.0]$ m.
The problem constraints $g, h$ include the terminal constraint for goal-reaching and the collision avoidance constraint with the obstacles.

The end effector dynamics are modeled by a linear system:
\begin{align}
    \dot{p}_x = u_x, \quad \dot{p}_y = u_y, \nonumber
\end{align}
where $(p_x, p_y)$ denotes the planar position and $(u_x, u_y)$ denotes the control inputs.
The trajectory is discretized into 80 time steps. The decision variable is defined as $x = (t, u_x^1, u_y^1, \ldots, u_x^{80}, u_y^{80})$.

We use SNOPT to solve the trajectory optimization and collect solutions. 
A total of 237k locally optimal (feasible) trajectories were collected across 2,700 different problem instances $\mathcal{P}_y$, and evaluation is conducted on problems with previously unseen parameter values $y$.

\subsubsection{Two-Car Reach-Avoid}

This problem aims to plan time-optimal trajectories for two cars, each navigating to its own goal location while avoiding collisions with each other and with randomly placed obstacles, as shown on the right of Fig.~\ref{fig: example trajectory}.

The problem parameters are given by $y = (p_{\text{obs}}, r_{\text{obs}})$, where two obstacles are randomly positioned between the start and goal locations. 
The obstacle radii are sampled from $[0.5, 1.5]$ m.
The problem constraints $g, h$ include the terminal constraint for goal-reaching and collision avoidance constraints between each car and the obstacles.

Each car follows a nonlinear dynamic model:
\begin{align}
    \dot{p}_x = v \cos \theta, \quad \dot{p}_y = v \sin \theta, \quad \dot{v} = a, \quad \dot{\theta} = \omega, \nonumber
\end{align}
where $(p_x, p_y)$ denote the position, $v$ denotes the speed, $\theta$ denotes the orientation, and $(a, \omega)$ represent the control inputs for acceleration and angular velocity.
Each trajectory is discretized over 40 time steps. 
The optimization variable is defined as $x = (t, a_1^1, \omega_1^1, a_2^1, \omega_2^1, \ldots, a_1^{40}, \omega_1^{40}, a_2^{40}, \omega_2^{40})$,
where the subscripts 1 and 2 denote the two cars.

Using SNOPT, we generate 114k locally optimal and feasible trajectories from 3,000 different problem instances $ \mathcal{P}_y$. 
Evaluation is again performed on unseen configurations of $y$.

\subsection{Training setup}

\begin{figure}[tp]
    \centering
    \includegraphics[width=0.75\linewidth]{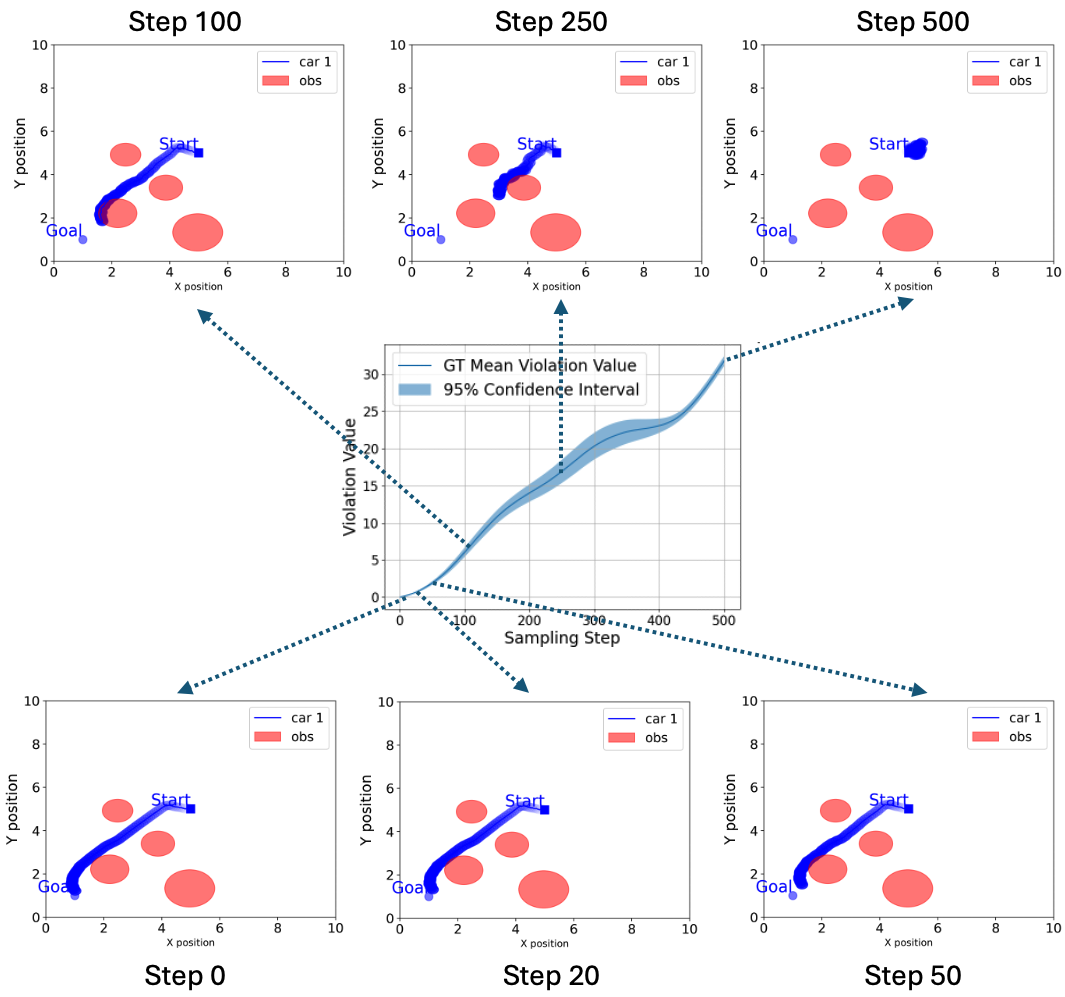}
    \caption{Tabletop Manipulation: gradually corrupted trajectories.}
    \label{fig:tabletop gradually corrupted trajectories. Total sampling step 500.}
\end{figure}

\begin{figure}[tp]
    \centering
    \includegraphics[width=0.75\linewidth]{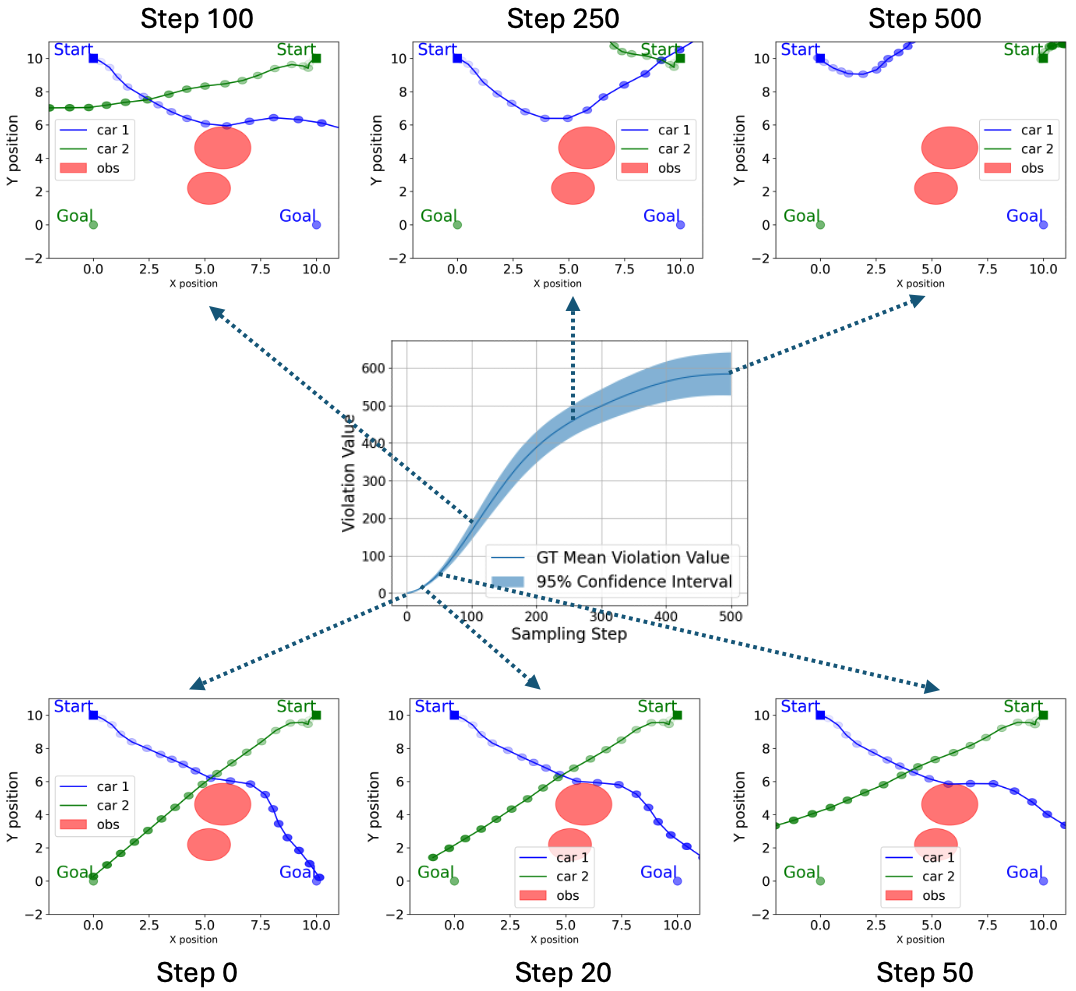}
    \caption{Two-car Reach-Avoid: gradually corrupted trajectories.}
    \label{fig:car gradually corrupted trajectories. Total sampling step 500.}
\end{figure}

For each problem, we train both the unconstrained diffusion model and the constraint-aligned diffusion model using 500 diffusion steps for training and testing. 
Both models use a U-Net architecture \cite{ronneberger2015u} with three hidden layers consisting of 512, 512, and 1024 neurons, along with fully connected layers to encode conditional information using 2 hidden layers of 256 and 512 neurons.

All models are trained using the Adam optimizer \cite{kingma2014adam} with 3 random seeds, each for 200 epochs. 
Training takes 3 to 8 hours for vanilla unconstrained diffusion models and 14 to 55 hours for constraint-aligned diffusion models across the two problems. 
This substantial difference in training time reflects the computational complexity introduced by incorporating the constraint violation loss. 
Note that, during inference, both model have equivalent sampling time since they share identical architecture, differing only in their learned weights resulting from the distinct loss functions used in training.

\subsection{Ground Truth Violation Analysis}

In Fig. \ref{fig: tabletop car gt violation}, we illustrate how ground truth constraint violation statistics evolve through 500 diffusion steps as we gradually corrupt the training data for both the tabletop manipulation and two-car reach-avoid problems. To generate these plots showing mean and 95\% confidence intervals of violation values, we sample 128 data points from the training data and corrupt each with 100 random samples at each diffusion step using forward sampling in Eq. \eqref{eq: forward sample}.
Our analysis reveals that ground truth violations in both problems increase as data becomes noisier in later steps,
though the rate of increase and magnitude of values differ significantly between problems. 

Figs. \ref{fig:tabletop gradually corrupted trajectories. Total sampling step 500.} and \ref{fig:car gradually corrupted trajectories. Total sampling step 500.} visualize these gradually corrupted trajectories alongside their corresponding ground truth violation values for both problems.
From the plot, we observe that in the early sampling steps---when the data is less noisy---constraint violations are primarily caused by both obstacle collisions and failure to reach the goal. 
In later steps, the violations are predominantly due to not satisfying the terminal goal-reaching constraint.
This analysis provides a ``standard" for expected constraint satisfaction at each diffusion step. 
During training, we leverage this information to align the constraint violation of predicted samples $\mathcal{L}_\text{vio}(x_0, y, k)$ with the ground truth violation $\mathcal{\mu}_{\text{vio\_GT}} (x_0, y, k)$ through our re-weighted constraint violation loss in Eq. \eqref{eq: hybrid loss}.

\subsection{Trajectory Sample Evaluation}

\begin{figure}
    \centering
    \includegraphics[width=1.0\linewidth]{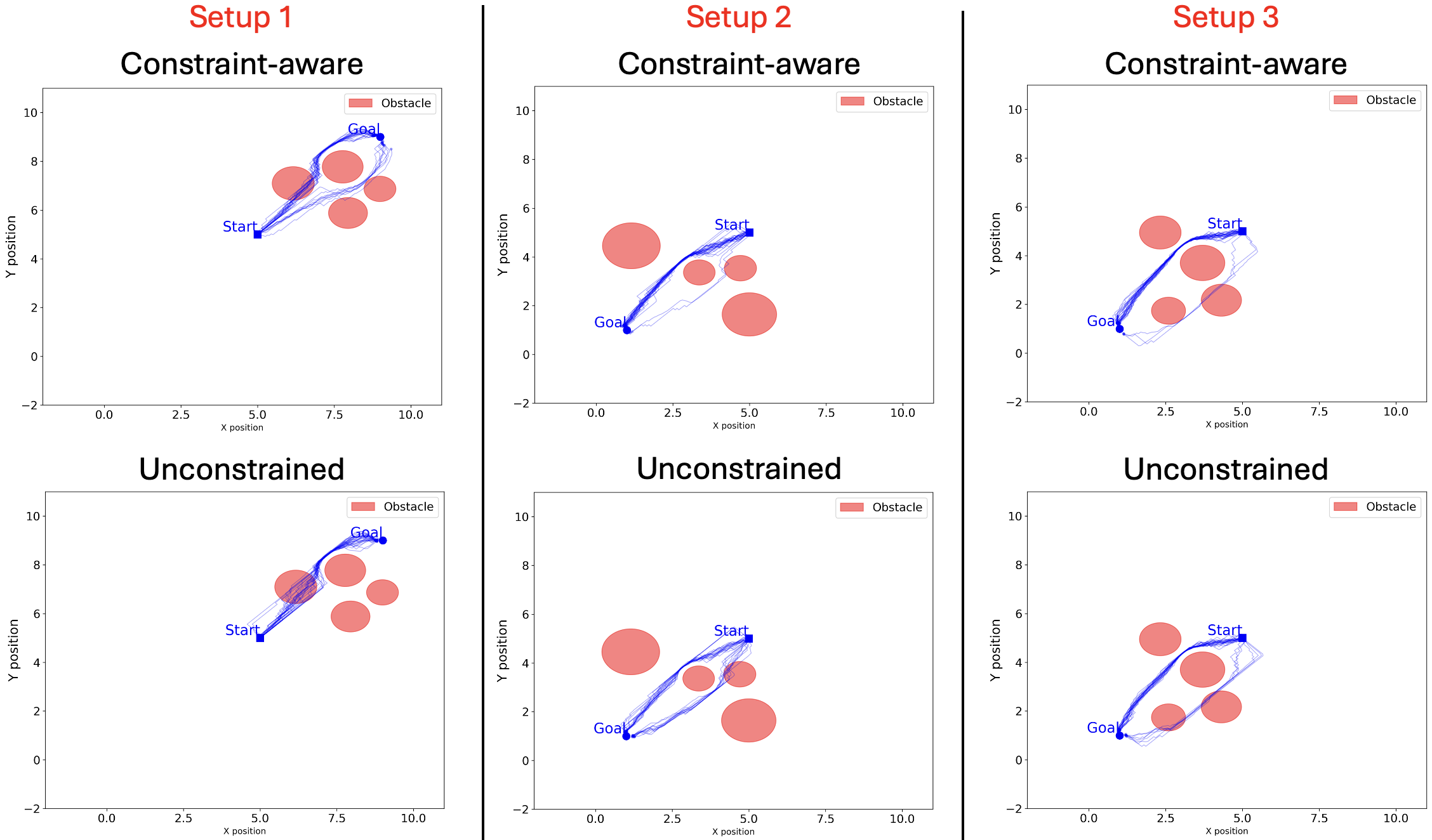}
    \caption{Trajectory samples for tabletop manipulation problem.}
    \label{fig:trajectory distribution for tabletop}
\end{figure}

In Fig. \ref{fig:trajectory distribution for tabletop}, we visualize trajectory samples from both constraint-aligned diffusion and unconstrained diffusion models across three different tabletop manipulation setups. 
Both models successfully sample from multimodal distributions conditioned on problem parameters such as goal and obstacle locations. 
However, our constraint-aligned diffusion model consistently generates more trajectory samples in feasible regions.
For example, in setups 1 and 2, our constraint-aligned diffusion model identifies that there is an upper route and a lower route, and one of them has a wider opening.
Consequently, our model generates more samples along the wider route, demonstrating the symmetry-breaking behavior in optimal control theory \cite{kappen2005path}.
The unconstrained diffusion model fails to make this distinction. 
This behavior demonstrates how the constraint-aligned diffusion model effectively shifts the sampling distribution toward feasible regions through the hybrid loss function during training.

\begin{figure}
    \centering
    \includegraphics[width=1.0\linewidth]{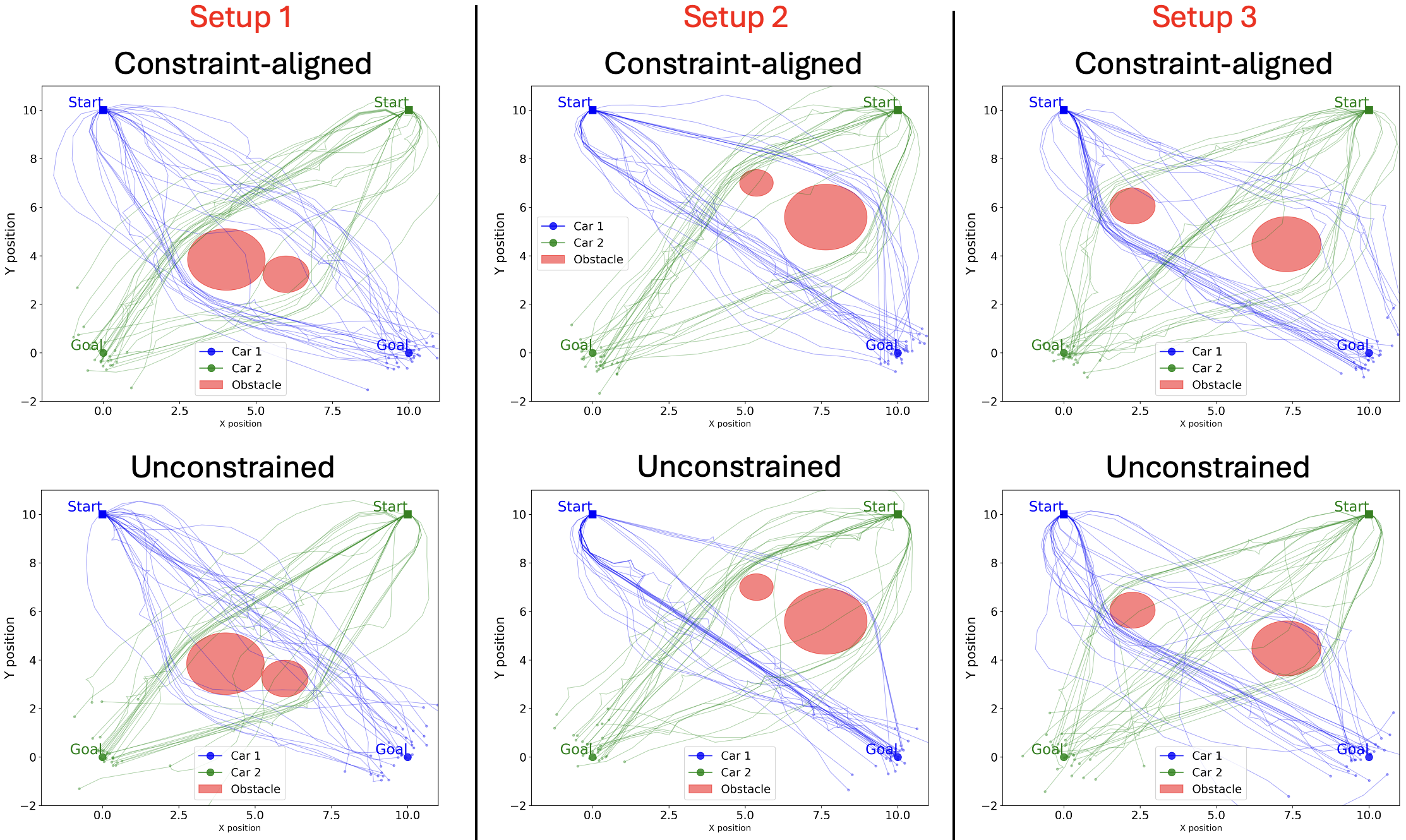}
    \caption{Trajectory samples for two-car reach-avoid problem.}
    \label{fig:trajectory distribution for car}
\end{figure}

In Fig. \ref{fig:trajectory distribution for car}, we present trajectory samples from both models across three different two-car reach-avoid scenarios. Our constraint-aligned diffusion model consistently produces paths with fewer collisions with the red obstacles compared to the unconstrained diffusion model. Furthermore, trajectory samples from our constraint-aligned model reach the designated goal locations more accurately, indicating better satisfaction with the goal-reaching constraints.

In Table \ref{table:constraint_violation}, we present the constraint violation statistics and feasibility ratios measured across model samples. 
For both problems, our constraint-aligned diffusion model achieves significantly lower constraint violations in the top 25\% of samples and produces more completely feasible solutions (with zero violations) compared to the unconstrained diffusion model. 
As a baseline reference, uniform random sampling of control variables typically results in high constraint violations.

We additionally demonstrate that our constraint-aligned diffusion model maintains solution optimality while reducing constraint violations. 
To verify this, we feed samples from both constraint-aligned and unconstrained diffusion models to the SNOPT solver as initial guesses and measure the computational time required to converge to locally optimal solutions. 
The shorter the convergence time, the closer the initial guess is to a local optimum.
Table \ref{table:solution_solving_time} presents the computational time statistics required for SNOPT to converge across both methods and uniform random initial guesses. 
The results show that the constraint-aligned diffusion model obtains locally optimal (and thus feasible) solutions with a time similar to that of the unconstrained diffusion model, both significantly faster than uniform sampling. 
This demonstrates that our proposed constraint-aligned diffusion model successfully improves constraint satisfaction while preserving the ability to effectively sample approximately local optimal solutions.

\begin{table}[t]
\caption{Constraint violation statistics and feasible ratio of samples.}
\label{table:constraint_violation}
\vskip 0.15in
\begin{center}
\begin{small}
\begin{sc}
\begin{tabular}{llccc}
\toprule
Problem & Method & Mean($\pm$STD) & 25\%-Quantile & Feasible Ratio \\
\midrule
\multirow{2}{*}{Tabletop} & Constr. Diff.      & 4.80 $\pm$ 5.61 & \textbf{0.04} & \textbf{58.3\textperthousand} \\
                          & Diffusion          & \textbf{4.73 $\pm$ 5.73} & 0.11 & 8.5\textperthousand \\
                          & Uniform            & 32.20 $\pm$ 3.73 & 29.66 & 0\textperthousand\\
\midrule
\multirow{2}{*}{Two-car} & Constr. Diff.      & \textbf{2.72 $\pm$ 20.30} & \textbf{0.54} & \textbf{0.4\textperthousand} \\
                         & Diffusion          & 10.99 $\pm$ 35.59 & 1.98 & 0\textperthousand \\
                         & Uniform            & 545.66 $\pm$ 294.71 & 353.96 & 0\textperthousand \\
\bottomrule
\end{tabular}
\end{sc}
\end{small}
\end{center}
\vskip -0.1in
\end{table}

\begin{table}[t]
\caption{Computational time required for obtaining locally optimal solutions when warm-starting solvers with generated samples as initial guesses.}
\label{table:solution_solving_time}
\vskip 0.15in
\begin{center}
\begin{small}
\begin{sc}
\begin{tabular}{llcccr}
\toprule
Problem & Method    & Mean($\pm$STD) & 25\%-Quantile & Median \\
\midrule
\multirow{2}{*}{Tabletop} & Constr. Diff.   & 7.63 $\pm$ 16.99 & 1.06 & \textbf{1.36} \\
                          & Diffusion           & \textbf{7.51 $\pm$ 17.50} & \textbf{0.73} & 1.37 \\
                          & Uniform            & 31.65 $\pm$ 24.55 & 14.81 & 22.72 \\
\midrule
\multirow{2}{*}{Two-car} & Constr. Diff.    & 19.32 $\pm$ 14.70 & 8.99 & \textbf{15.33} \\
                         & Diffusion           & \textbf{18.82 $\pm$ 14.33} & \textbf{8.77} & 15.61 \\
                         & Uniform             & 46.17 $\pm$ 20.63 & 29.62 & 43.54 \\
\bottomrule
\end{tabular}
\end{sc}
\end{small}
\end{center}
\vskip -0.1in
\end{table}

\section{Conclusions}

In this paper, we introduced a novel approach that explicitly aligns diffusion models with problem-specific constraint information to enhance trajectory optimization. 
With statistical analysis of ground truth constraint violations, we allow close alignment between diffusion samples and ground truth violations through a re-weighting scheme in the diffusion process.
Demonstrated on tabletop manipulation and two-car reach-avoid problems, our approach effectively reduces constraint violations while generating high-quality trajectory solutions. 

While our method achieves strong performance, it currently incurs notable computational overhead, primarily due to the use of numerical integration and extensive sampling required to statistically characterize constraint violations during training.
Future work could explore more efficient model architectures and training pipelines tailored to constraint-aligned diffusion models, potentially reducing computational demands without compromising performance. 
Additionally, incorporating real-time environmental feedback and developing iterative fine-tuning strategies could further enhance the model's adaptability to dynamic environments.
These directions offer promising opportunities for making constraint-aligned diffusion models even more efficient, robust, and applicable in dynamic, real-world trajectory optimization.


%
%
\bibliographystyle{splncs04}
\bibliography{main.bbl}

\end{document}